\documentclass[10pt,twocolumn,letterpaper]{article}

\usepackage[pagenumbers]{cvpr} 

\usepackage[accsupp]{axessibility}  
\usepackage{graphicx}
\usepackage{amsmath}
\usepackage{amssymb}
\usepackage{booktabs}
\usepackage{multirow}
\usepackage{bm}
\usepackage{color}
\newenvironment{packed_itemize}{
\begin{itemize}
  \setlength{\itemsep}{1pt}
  \setlength{\parskip}{0pt}
  \setlength{\parsep}{0pt}
}{\end{itemize}}
\usepackage[ruled,linesnumbered]{algorithm2e}
\DeclareMathOperator*{\argmax}{argmax}

\usepackage[pagebackref,breaklinks,colorlinks]{hyperref}

\usepackage[capitalize]{cleveref}
\crefname{section}{Sec.}{Secs.}
\Crefname{section}{Section}{Sections}
\Crefname{table}{Table}{Tables}
\crefname{table}{Tab.}{Tabs.}

\def\cvprPaperID{11777} 
\def\confName{CVPR}
\def\confYear{2022}

\begin{document}

\title{Fine-Grained Object Classification via Self-Supervised Pose Alignment}

\author{Xuhui Yang$^1$, Yaowei Wang$^{1*}$, Ke Chen$^{2,1}\thanks{Corresponding authors}$~, Yong Xu$^{1,2,3}$, Yonghong Tian$^1$ \\
$^1$Peng Cheng Laboratory $^2$South China University of Technology \\
$^3$China Communication and Computer Network Laboratory of Guangdong\\
{\tt\small \{yangxh, wangyw\}@pcl.ac.cn, \{chenk, yxu\}@scut.edu.cn, tianyh@pcl.ac.cn}
}

\maketitle

\begin{abstract}
Semantic patterns of fine-grained objects are determined by subtle appearance difference of local parts, which thus inspires a number of part-based methods.
However, due to uncontrollable object poses in images, distinctive details carried by local regions can be spatially distributed or even self-occluded, leading to a large variation on object representation. 
For discounting pose variations, this paper proposes to learn a novel graph based object representation to reveal a global configuration of local parts for self-supervised pose alignment across classes, which is employed as an auxiliary feature regularization on a deep representation learning network.
Moreover, a coarse-to-fine supervision together with the proposed pose-insensitive constraint on shallow-to-deep sub-networks encourages discriminative features in a curriculum learning manner.  
We evaluate our method on three popular fine-grained object classification benchmarks, consistently achieving the state-of-the-art performance. 
Source codes are available at \url{https://github.com/yangxh11/P2P-Net}.

\end{abstract}

\section{Introduction}
Semantic patterns in image-based object classification are determined by visual appearance and shape of object classes.
The problem of classifying fine-grained objects is made more challenging due to the inherently subtle shape difference across the subordinate categories (\eg bird breeds \cite{birds2011} and vehicle brands \cite{cars2013}). 
As a result, fine-grained classification relies on distinguished appearance details on local parts, which typically desires detection on those specific parts additionally using expensive part annotations \cite{lac2015, mask2018}.  
Learning a discriminative representation for fine-grained objects remains non-trivial in the context of deep learning.
Generally, a good representation in fine-grained classification should not only be sensitive to the subtle detail changes that usually anchored on specific parts but also be invariant to the deformations of object parts and the changes of viewing angles.


On the one hand, a large number of deep approaches were devoted to the former challenge of capturing fine-grained details in local regions, which relied on detecting discriminative parts to extract local features to complement the global one \cite{TASN2019, ASD2020, NTSNET2018}.
On the other hand, another group of deep methods \cite{BCNN-lowrank2017, BCNN-attention2018, BCNN-group2019} avoided explicitly localizing object parts but tried to enhance the object representation's discrimination by applying bilinear pooling operation on feature vectors from two independent network streams.
However, only a few works have paid attention to the challenge of feature inconsistency caused by the changes of object pose or viewing angle.

\begin{figure}[t]
\centering
\includegraphics[width=0.9\linewidth]{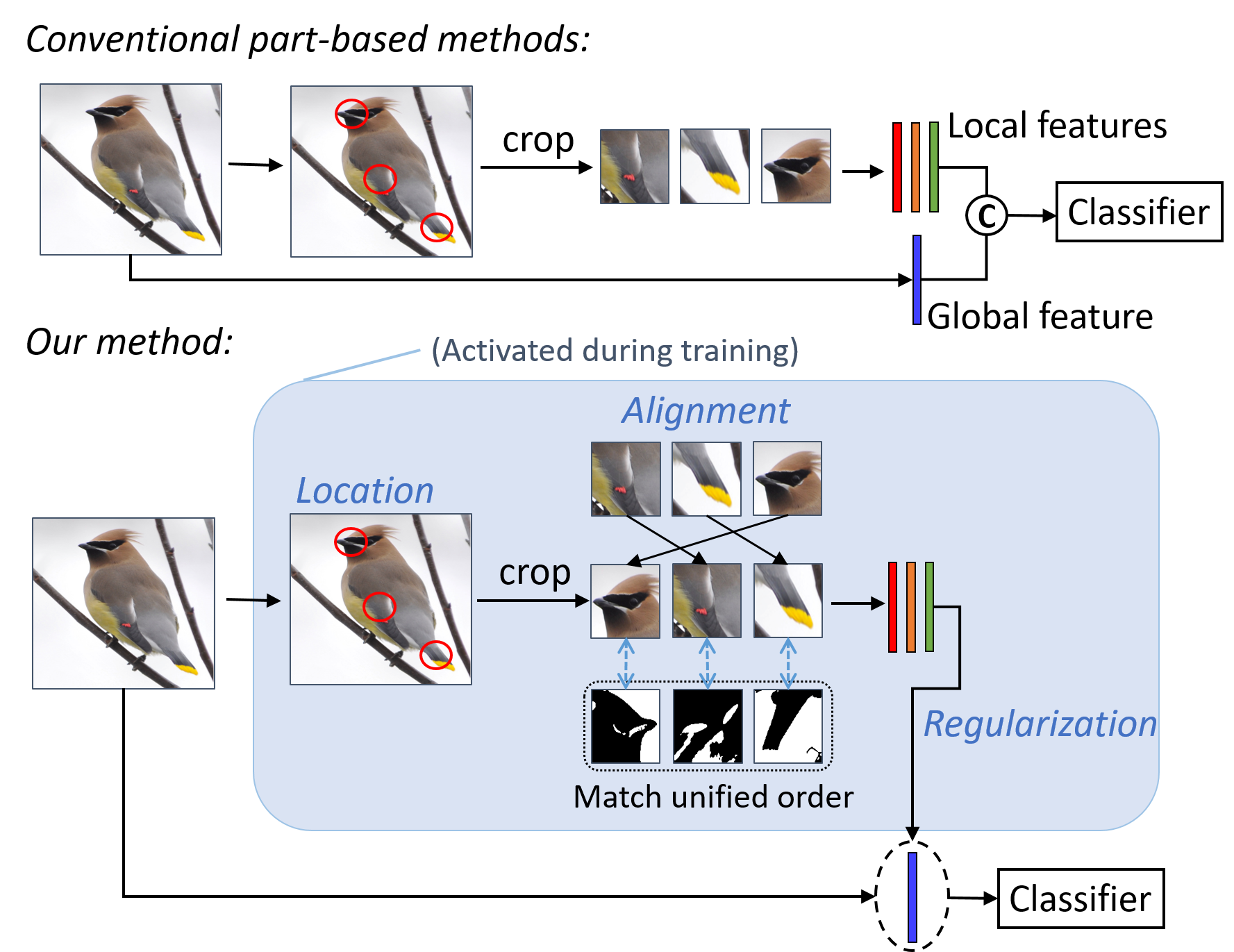}
\caption{
Different from direct concatenation of local and global features in conventional part-based methods, the main idea of the proposed P2P-Net is to incorporate a pose-insensitive configuration of local distinctive details into object representation via utilizing self-supervised pose alignment as feature regularization to narrow intra-category variance and enlarge inter-category margin.
}
\label{fig:motivation} \vspace{-0.5cm}
\end{figure}

Similar to other pose estimation problems \cite{openpose2019,vehicle_pose2020}, poses of fine-grained objects can approximately be described as a geometric configuration of discrete object parts. 
As a result, we argue that \textit{reliable object parts localization} and \textit{parts alignment} are essential to model object poses, which can be utilized to generate pose insensitive object representations \cite{pose-norm2020} so as to eliminate object pose variations.
Evidently, explicitly regression on object poses as an auxiliary task is a straightforward solution, but in the context of fine-grained classification object poses typically suffers from inherent annotation ambiguities and a lack of sufficient samples. 
In view of this, we propose an unsupervised part-to-pose network (P2P-Net) which captures both part details and object poses to regularize the representation learning with no additional annotations of part position or pose information.
Inspired by Feature Pyramid Networks (FPN) \cite{fpn2017} that is originally designed for object detection, our P2P-Net distinguishes the confidently class-discriminative regions from region proposals in a weakly supervised learning manner. 
To incorporate the fine-grained appearance details on specific parts, the global object representation is regularized with local representations of the detected salient regions via minimizing distribution difference with the contrastive loss.

The major differences between the P2P-Net and the existing part-based methods, as shown in \cref{fig:motivation}, lie in: 1) we align the parts in a self-supervised learning manner so as to discount the pose variations; 2) a feature regularization to strengthen representation discrimination is available only at training, \ie, our method can discard the time consuming part branch (highlighted in light blue in \cref{fig:motivation}) during testing. 
The effectiveness of the proposed self-supervised part-based pose alignment as feature regularization can be demonstrated in our experiments.
Besides, we also propose curriculum supervision to capture discriminative details at coarse-to-fine scales for further improvement.
The proposed P2P-Net consistently outperforms the state-of-the-art fine-grained classifiers on three popular benchmarks.

Our contributions can be summarized as follows:
\begin{packed_itemize}
\item This paper proposes an end-to-end P2P-Net to incorporate discriminative features on confidential parts and then promotes the discrimination of object representations via pose-insensitive feature regularization. 
\item Technically, our P2P-Net designs an adaptive graph matching algorithm on confidential parts and achieves feature consistency against pose variations in an unsupervised learning manner.
\item A generic curriculum supervision on image classification is introduced by designing label-smoothing based easy-to-hard supervision signals coupled with shallow-to-deep sub-networks. 
\item Experimental results on multiple public benchmarks demonstrate that the proposed method can achieve a new state-of-the-art performance on the problem of fine-grained image classification .
\end{packed_itemize}

\section{Related Work}\label{sec:related_work}
In the past decades, deep learning has made remarkable progresses in the application of fine-grained image classification \cite{hierarchical2018, mcloss2020}. 
These methods can be categorized into two main groups --  part-free and attention/part-based methods.

The former group of algorithms emerges a trend that a part of fine-grained recognition methods seek to boost the backbone's recognition capability by either weighting features using attention maps or employing augmented samples as training data.
Lin \etal proposed B-CNN~\cite{BCNN2015}, which uses two-stream architectures to simultaneously model the location and appearance of details.
Zheng \etal \cite{MACNN2017} first explored to spot salient peaks by clustering spatially-correlated channels, and then in \cite{TASN2019} used a trilinear attention module to model the inter-channel relationships and then distorted the image via attention based pixel sampling. 
Ding \etal \cite{S3N2019} performed a selective sparse sampling operation to discover diverse and fine-grained details. 
Chen \etal \cite{DCL2019} first proposed an image splitting operation for fine-grained classification by partitioning the image into local patches, which are randomly shuffled to reconstruct a new image for training, and thus force the model to concern on local details rather than its global configuration.
Du \etal \cite{jigsaw2020} took one step further in applying a multi-scale jigsaw puzzle generator to capture cross-granularity information with a progressive training strategy.
Inspired by these part-free methods, we propose a curriculum training scheme generic to all image classification tasks, which can improve model generalization by using easy-to-hard supervision signals coupled with shallow-to-deep sub-networks. 

\begin{figure*}[t]
\centering
\includegraphics[width=0.84\linewidth]{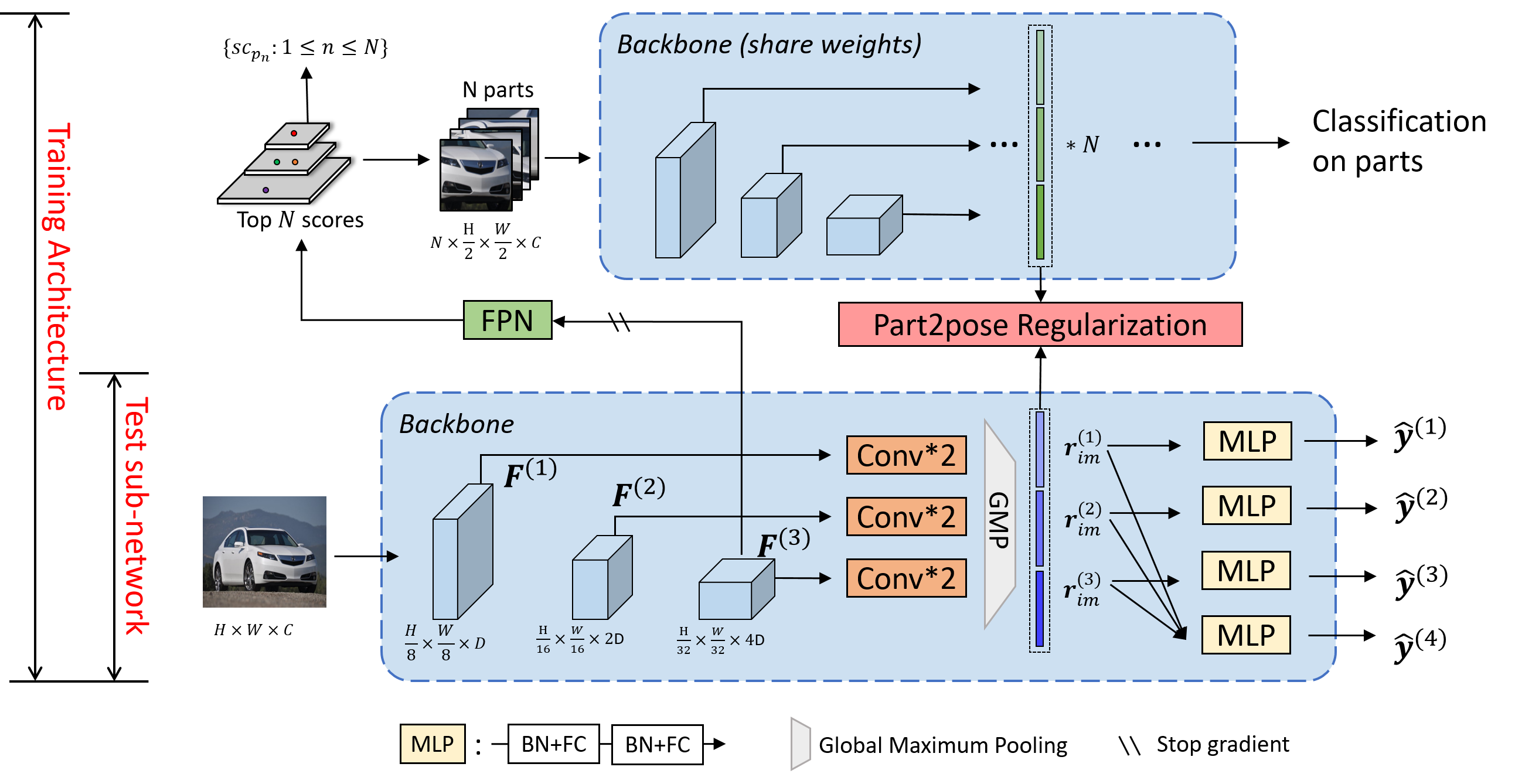}
\caption{Pipeline of the proposed P2P-Net. 
Both global feature encoding backbone (bottom) and part2pose feature regularization (top) are utilized 
for training while only the bottom one is activated during testing.
Network parameters of the feature encoders (highlighted in light blue) in both upper and bottom branches are shared. 
$\bm{F}^{(s)}$ denotes an intermediate feature map from convolutional layer at specific depth, where $1 \leqslant s \leqslant S$ and $S=3$, while $\bm{\hat{y}}^{(s)}$ denotes classification predictions made on representations $\bm{r}_{im}^{(s)}$. 
Note that, $\bm{\hat{y}}^{(S+1)}$ is made on the concatenation of $\{\bm{r}_{im}^{(s)}\}$.
For the detected parts, $sc_{p_n}$ denotes the score of the corresponding part $p_n$, where $1\leqslant n \leqslant N$.
Detailed structure of the feature pyramid network (FPN) block and part2pose regularization block are shown in \cref{fig:fpn}.
}
\label{fig:architecture}\vspace{-0.5cm}
\end{figure*}

It is observed that capturing the subtle differences in discriminative parts plays an important role to distinguish similar categories.
In light of this, a number of attention/part-based methods have been encouraged to detect and learn discriminative features according to the attention/score maps or region proposals \cite{focus_parts2019, complementary_parts2019, twolevel2015}. 
Early works detected parts based on densely part annotations \cite{part_rcnn2014,lac2015} in a fully supervised learning style.
Recent studies relaxed the desire of expensive part annotations and localized informative regions in a weakly-supervised way, \ie, only given category labels.
In \cite{RACNN2017}, a recurrent attention convolutional neural network is proposed to iteratively zoom in local discriminative regions and introduced an inter-scale ranking loss between multi-scale regions to reinforce feature learning. 
Wang \etal \cite{CDL2019} proposed a discriminative region grouping sub-network to discover discriminative regions, while Yang \etal \cite{NTSNET2018} introduced a weakly supervised feature pyramid network into fine-grained classification to select top distinguishable parts.
Other works further investigated the spatial or context relations among object parts \cite{CDL2019, part_constraint2017, spatial_relation2019}.
Those part-based methods are robust on locating discriminative parts, but usually fused features of parts and image with a simple concatenation without further exploiting the latent correlation across parts features \cite{mixture2019, wsdan2019}, and thereby inevitably encountered the feature alignment problem.
Our approach follows the same way of weakly supervised part-based approaches, but we go three steps further to learn pose insensitive representations: 1) aligning parts according to their correlations; 2) enhancing the discrimination of object representation with detected parts via feature regularization instead of feature concatenation; 3) encouraging discriminative features in a curriculum learning manner. 


\section{Methodology}\label{sec:methodology}

For the problem of fine-grained image classification with $\mathbb{X}$ and $\mathbb{Y}$ denoting the input and output space, given $L$ training samples $\{\bm{I},y\}^L$, where $\bm{I}\in \mathbb{X}$ and  $y \in \mathbb{Y}$ denote one visual observation (\ie image) and its corresponding category label, the goal is to learn a mapping function  $\Phi: \mathbb{X} \rightarrow \mathbb{Y}$ that correctly classifies images into one of the $|\mathbb{Y}| = K$ categories.
In this section, we present the whole architecture of our P2P-Net, as shown in \cref{fig:architecture}, which is composed of three parts: 
1) curriculum supervision for complementary representation learning (see \cref{sec:curri_learning});
2) contrastive feature regularization for strengthening object representation's discrimination (see \cref{sec:parts_regularization}); 
3) unsupervised graph matching method for part alignment (see \cref{sec:parts_alignment}).
The testing sub-network is only a part of the whole architecture, 
which greatly reduces the computational complexity during inference since only the bottom backbone is activated.

\subsection{Curriculum Supervision on Backbone Network} \label{sec:curri_learning}
In the problem of fine-grained image classification, a typical ResNet~\cite{resnet2016} (\eg, ResNet34 or ResNet50) is adopted as backbone in this paper.
As \cref{fig:architecture} shows, several intermediate feature maps of the backbone at different depths are fed into different convolutional blocks respectively.
They are noted as $\bm{F}^{(s)}$, where $s\in \{1,...,S\}$ stands for different stages and $s$ is proportional to the sub-network depths. 
Each block is followed by a global maximum pooling (GMP) layer to obtain image representations $\{\bm{r}_{im}^{(s)}\}$.
At the final layer we stack several independent multi-layer perceptrons (MLP) for classification separately. 
As a result, there are in total $S+1$ predictions for one image, which will be aggregated for the final prediction, whose detail is introduced in \cref{sec:training_inference}.
Note that, the above design of training sub-networks is generic and can readily be applied to existing classification models.

Beyond employing a conventional ResNet as backbone, we introduce a curriculum training scheme encouraged by \cite{jigsaw2020}, which is based on the observation that deeper network has stronger learning capability and tends to correctly distinguishes more challenging samples.
To improve model generalization and encourage diversity of representations from specific layers, we introduce a curriculum training strategy by employing a soft supervision method based on label smoothing \cite{label_smooth2019}. 
The cross-granularity representations can then be utilized either independently or jointly to make better predictions. 
We modify the one-hot vector as follows:
\begin{equation}
    \bm{y}_{\alpha}[t]=
    \begin{cases}
    \alpha, & t = y \\
    \frac{1-\alpha}{K}, & t \neq y
    \end{cases}
    ,
\end{equation}
where $\alpha$ denotes a smoothing factor between 0 and 1, $t$ represents the element index of the label vector $\bm{y} \in \mathbb{R}^K$. 
Note that, $\alpha$ controls the magnitude of the ground truth class in the new target $\bm{y}_{\alpha}$, which can thus be used to construct easy-to-hard curriculum targets coupled with shallow-to-deep feature encoders.
As a result, predictions $\{\bm{\hat{y}}^{(s)}\}$ on representations from different layers or their combination are supervised with different labels during training, whose loss function can be written in the following form:
\begin{equation}\label{eq:sce}
\begin{aligned}
    \ell_{sce}(\bm{\hat{y}}^{(s)}, y, \alpha^{(s)}) 
    & = \ell_{ce}(\bm{\hat{y}}^{(s)}, \bm{y}_{\alpha^{(s)}}) \\
    & = \sum^{K-1}_{t=0}{-\bm{y}_{\alpha^{(s)}}[t] \log(\bm{\hat{y}}^{(s)}[t])}, \\
\end{aligned}
\end{equation}
where $\ell_{sce}(\cdot)$ denotes a smooth cross-entropy loss which takes one more smoothing factor than the normal cross-entropy loss \cite{ml_prob2012}.
Since $S+1$ predictions are available in our P2P-Net, the overall classification loss of each image sample can be written as follows: 
\begin{equation} \label{eqn:im_cls}
    L_{im}^{cls} = \sum_{s=1}^{S+1} {\ell_{sce}(\bm{\hat{y}}^{(s)}, y, \alpha^{(s)})}.
\end{equation}
Generally, deeper sub-networks have more capacity to learn a mapping from more challenging samples. 
Therefore, we gradually increase $\alpha^{(s)}$ from a value greater than $\frac{1}{K}$ to 1.
As $s$ getting larger, the supervision label $\bm{y}_{\alpha^{(s)}}$ approaches more closer to the one-hot target code, which indicates that the corresponding sub-network should make more confident prediction with deeper network architecture.


\subsection{Contrastive Feature Regularization} \label{sec:parts_regularization}

\begin{figure}[t]
\centering
\includegraphics[width=0.98\linewidth]{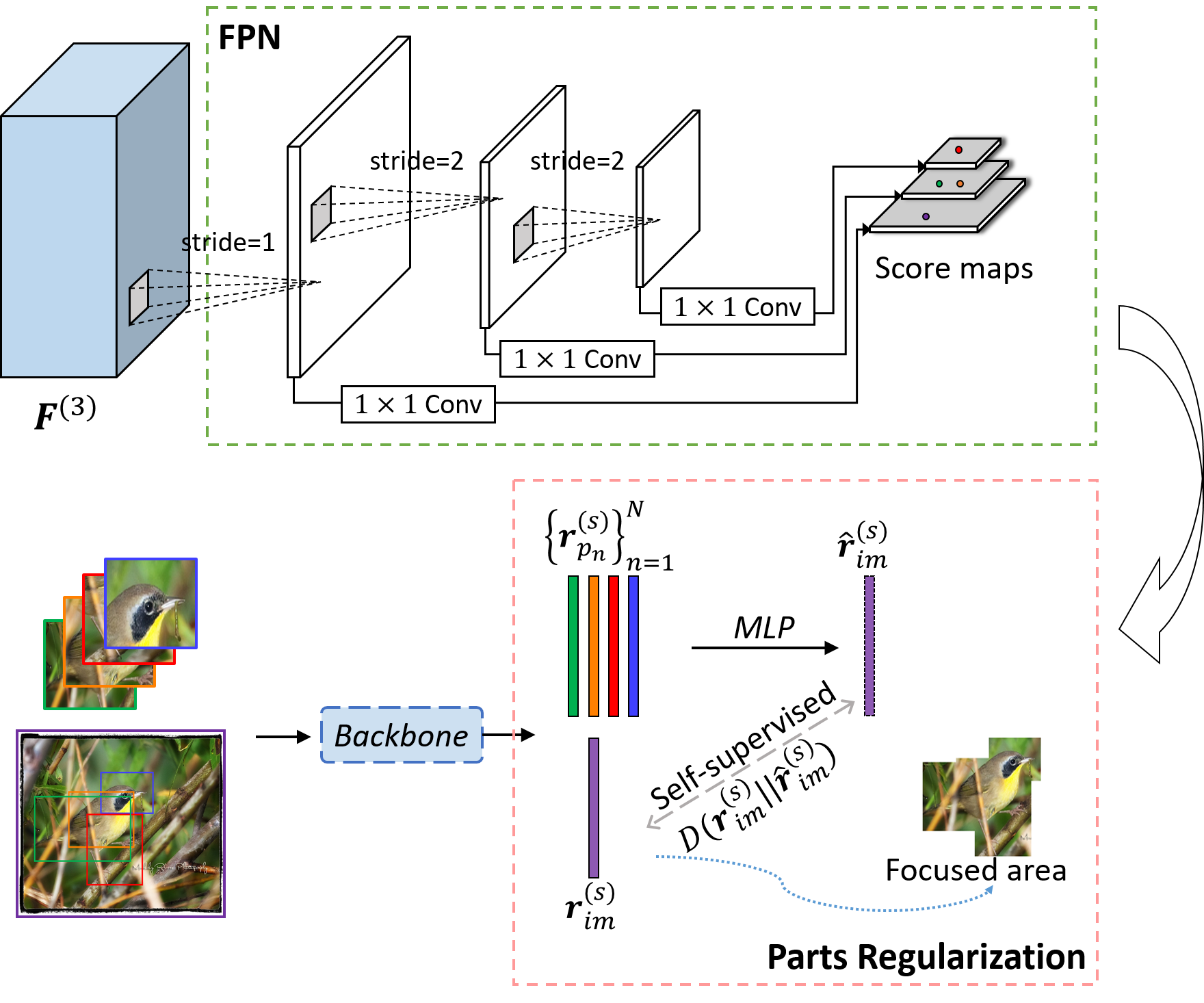}
\caption{Illustration of the FPN and the part-based contrastive feature regularization. As shown in \cref{fig:architecture}, the input of FPN is $F^{(3)}$ from the backbone while the outputs are score maps of different scales. 
We apply the non-maximum suppression (NMS) to the score maps and select the top $N$ discriminative parts, features from which are then used to regularize the object representation learning in a self-supervised contrastive learning manner.
}
\label{fig:fpn}\vspace{-0.5cm}
\end{figure}

Locating discriminative parts is an effective way to mitigate the challenge of inter-class similarity.
Typically, in existing works \cite{MACNN2017,DFGMM2020},  representations of parts are concatenated or fused to generate a new representation (to complement the global object representation) for classification.
In this paper, we treat the parts localization problem as an object detection task and propose a feature regularization on representations between local parts and global object to enforce incorporating fine-grained details from distinctive parts into image representation. 

\vspace{0.1cm}
\noindent\textbf{Weakly-Supervised Part Localization --} 
As shown in \cref{fig:fpn}, the backbone's last feature block is followed by a feature pyramid network (FPN), which generates a pyramid of score maps of different spatial sizes, \eg, $14\times14$, $7\times7$ and $4\times4$, by following the same setting as ~\cite{NTSNET2018}.
Each score element in the maps corresponds to a predefined image patch with a fixed size.
Note that, these patches having a unique size can be overlapped with others.
Moreover, elements in a larger map (\eg, $14\times14$) correspond to smaller image patches compared to the ones in a smaller map (\eg, $7\times7$).
Under the assumption that the scores indicate the discrimination degree of image patches, when selecting the top $N$ parts of highest scores, we apply the non-maximum suppression (NMS) to eliminate parts with large Intersection over Union (IoU).

\vspace{0.1cm} \noindent \textbf{Classification and Ranking Loss on Detected Parts --}
According to the top $N$ scores' indexes, we crop the corresponding image patches from the input image, which contain distinctive details of local object parts. 
For computational efficiency, these $N$ parts are then resized into $224 \times 224$ (half the spatial size of the original image) and fed into the backbone with shared weights.
The curriculum supervision described in \cref{sec:curri_learning} on the whole image is repeated for each resized patches (parts).
We define a part's feature as $\bm{r}_{p_n} = [\bm{r}^{(1)}_{p_n};\bm{r}^{(2)}_{p_n};\ldots;\bm{r}^{(S)}_{p_n}]$, where $1\leqslant n \leqslant N$.
Similar to \cref{eqn:im_cls}, the classification loss on the $n\text{-th}$ part can be depicted as
\begin{equation} \label{eqn:part_loss}
    L_{p_n} = \sum_{s=1}^{S+1} {\ell_{sce}(\bm{\hat{y}}_{p_n}^{(s)}, y, \alpha^{(s)})},
\end{equation}
and the total classification loss on all parts is
\begin{equation}
    L^{cls}_{parts} =  \sum^{N}_{n=1} L_{p_n}.
\end{equation}

Moreover, given top $N$ parts' classification losses $\{L_{p_n}\}^N$ and the corresponding scores $\{sc_{p_n}\}^N$,
we further argue that the prediction losses should be consistent with the confidence scores of image parts.
Given part indexes $n$ and $n'$, if $L_{p_{n}}<L_{p_{n'}}$, then $p_{n}$ should has a greater score than $p_{n'}$, so $sc_{p_{n}} > sc_{p_{n'}}$ is preferred in such a situation. 
As a result, we introduce an extra ranking loss based on a summation of conditional hinge losses \cite{hingeloss2004}:
\begin{equation} \label{eqn:ranking} 
\begin{aligned}
    L^{rank} 
    & = \sum^N_{n=1} \sum^N_{n'=1} \ell_{hg}(L_{p_{n}}, L_{p_{n'}}) * c_{nn'} \\
    & = \sum^N_{n=1} \sum^N_{n'=1} max(0, L_{p_{n}}-L_{p_{n'}}+\delta) * c_{nn'},
\end{aligned}
\end{equation}
with the indicator $c_{nn'}$ is defined as 
\begin{equation}
    c_{nn'}=
    \begin{cases}
    1, & sc_{p_{n}} > sc_{p_{n'}} \\
    0, & sc_{p_{n}} \leqslant sc_{p_{n'}}
    \end{cases}
    ,
\end{equation}
where $sc_{p_n}$ denotes the score of part $p_n$ and margin hyper-parameter $\delta$ is fixed as 1 in our paper.
The loss encourages $sc_{p_{n}}$ to be greater than $sc_{p_{n'}}$ if $L_{p_{n}}$ has a relative smaller value than $L_{p_{n'}}$.
Ideally, by minimizing the ranking loss, the scores $\{sc_{p_{n}}\}$ and the parts classification losses $\{L_{p_{n}}\}$ should change in an opposite way.
As a result, such a design can improve reliability of discriminative part detection.


\vspace{0.1cm}
\noindent \textbf{Contrastive Loss for Feature Regularization --}
Instead of making classification via concatenating local parts' and global image's features, we propose a feature regularization to constraint the object representation learning in a contrastive learning manner.
Given image representation 
$    \bm{r}_{im} = [\bm{r}^{(1)}_{im};\bm{r}^{(2)}_{im};...;\bm{r}^{(S)}_{im}]$
and part representation $\bm{r}_{p_n} = [\bm{r}^{(1)}_{p_n};\bm{r}^{(2)}_{p_n};...;\bm{r}^{(S)}_{p_n}]$,
we regularize each stage's representation with a contrastive loss as follows:
\begin{equation}
    \label{eq:reg_loss}
    L^{reg}=\sum_{s=1}^{S} \ell_{kl}(\bm{r}_{im}^{(s)}, \phi([\bm{r}_{p_1}^{(s)}; \bm{r}_{p_2}^{(s)}; ...; \bm{r}_{p_N}^{(s)}])),
\end{equation}
where $\ell_{kl}$ is the Kullback-Leibler divergence function, $\phi(\cdot)$ is an approximation function to be optimized, which we adopt a 2-Layer MLP to model it in the experiments.
Such a regularization loss can enforce object representation learning branch to focus on discriminative details from specific local regions, which is visually shown in \cref{fig:fpn}.
Through this way we can further filter out redundant misleading information in each $\bm{r}_{im}^{(s)}$ so as to improve discrimination of object representation.

\subsection{Graph Matching for Part Alignment} \label{sec:parts_alignment}

\begin{figure}[t]
\centering
\includegraphics[width=0.98\linewidth]{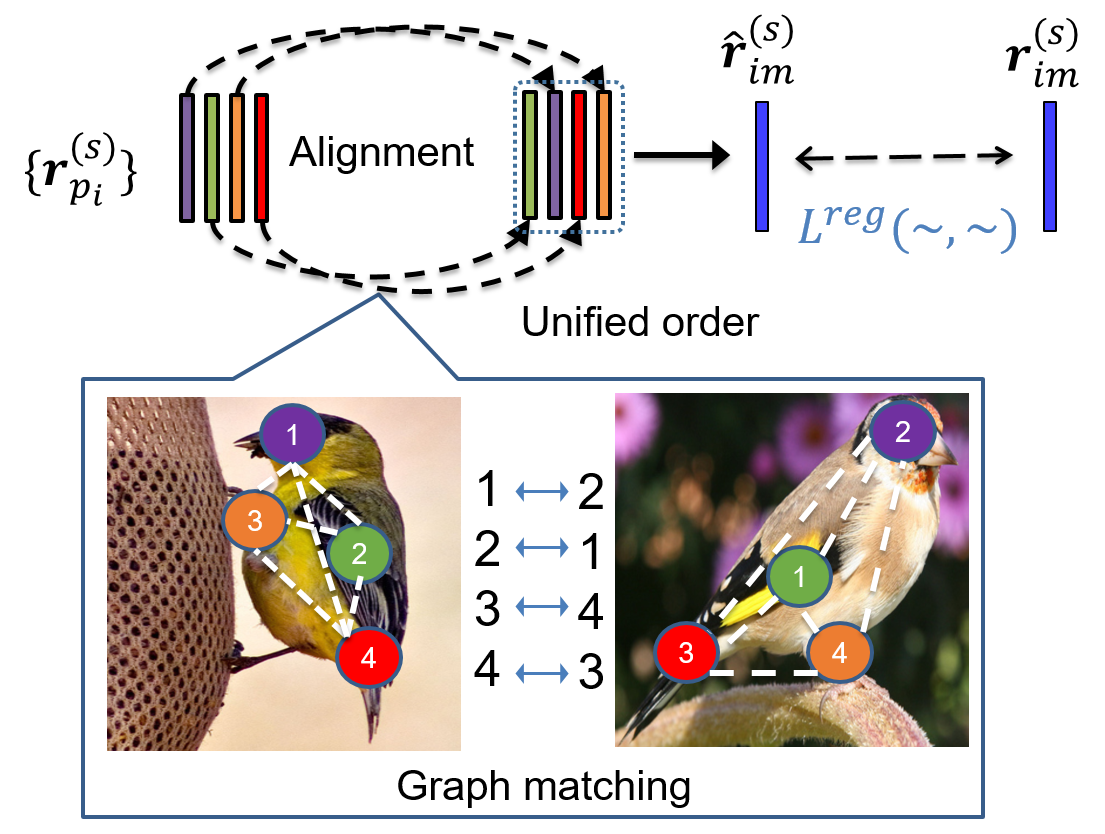}
\caption{Graph matching for part alignment.
For illustrative purpose, only one stage's representations are shown in the figure.
Located parts are regarded as ordered nodes according the discrimination scores.
We re-order the parts by applying graph matching on the parts correlation matrix according to its global configuration, whose features are then employed to regularized the learning of global image representation.
}
\label{fig:graph_matching}\vspace{-0.5cm}
\end{figure}

Although the top $N$ discriminative parts have been located through the \cref{sec:parts_regularization}, it is found out that the configuration of these parts is not consistent, \ie, they are not strictly aligned (See \cref{sec:visualization}).
As we only use category labels for supervision to discover discriminative parts, the exact semantic class of parts (\eg, head, body or tail) remains unknown. 
Therefore, simply concatenating the spotted parts' features in an arbitrary order for regularization may raise the feature inconsistency problem.
Fortunately, for any specific object category, its discriminative appearance usually appears on a limited size of local parts.  
Consequently, we propose an unsupervised graph matching method to sort the found parts in a unified order based on a basic assumption that the correlation between top $N$ parts is similar across images. 

The concept of our graph matching is illustrated in \cref{fig:graph_matching}.
Intuitively, it is supposed that top $N$ parts of the bird image on the left-hand side is ordered as $\textless head, wing, foot, tail\textgreater$ (\ie $\textless 1, 2, 3, 4\textgreater$) according to their scores $\{sc_{p_n}\}^4$, while the unified order maintained by our method may be $\textless wing, head, tail, foot\textgreater$.
The goal of our graph matching algorithm is to resort the order of detected object parts from $\textless 1, 2, 3, 4\textgreater$ to $\textless 2, 1, 4, 3\textgreater$ in the unified order for feature consistency.
In this way, resorted parts features discounting pose variations are used to constrain object representation by contrastive learning, which is formulated in \cref{eq:reg_loss}.

Technically, since we can not identify object's distinct parts directly, we maintain a unified correlation matrix to model latent relations in-between parts. 
The entry of the correlation matrix is given as follows:
\begin{equation} \label{eqn:corr_matrix}
    \bm{M}_{ij}=\textless\bm{r}_{p_i}, \bm{r}_{p_j}\textgreater,
\end{equation}
where $\bm{M}_{ij}$ denotes the relation score between part $p_i$ and part $p_j$.
Given a new image sample, we compute its parts correlation matrix, denoted as $\bm{M}'$, of each possible permutation of parts.
Then the $\bm{M}'$ has the largest matching degree with the reference matrix $\bm{M}$ is considered as the best alignment.
The formulation can be simplified as
\begin{equation}
   \hat{\bm{M}} = \mathop{\argmax}_{\bm{M}'}{vec(\bm{M}')}^{T}{vec(\bm{M})},
\end{equation}
where the matching degree can be simply measured by summarizing the element-wise product of vectorization of two matrices.

This is exactly a graph matching problem considering edges (relations) similarity but not nodes (parts) similarity.
In this paper, as parts also contain distinct classes information, it is not suitable to add nodes (parts) similarity even if they belong to the same type (\eg, head).
Note that we do not consider the mismatch situation in the experiments.
The permutation with the max matching degree is selected as the correct order and the algorithm returns the resorted parts representations.
Computation about graph matching is efficient because the size of parts $N\leqslant 5$ in this paper although there are $N!$ permutations in total.
Finally, for self-correction, an online-updating scheme is adopted to refine the parts centers and the reference matrix $\bm{M}$.
The parts centers are updated by weighting the new sample's resorted features and the old samples' parts features, which follows the basic rule that older samples are of smaller weights.

\subsection{Training and Inference} \label{sec:training_inference}

Given the aforementioned key components, our network can be effectively trained in an end-to-end manner. 
Specifically, the network loss is given as a summation of aforementioned losses with only one trade-off parameter $\beta$:
\begin{equation} \label{eqn:total_loss}
    L = L^{cls}_{im} + L^{cls}_{parts} + L^{rank} + \beta \cdot L^{reg},
\end{equation}
where $L^{cls}_{im}$, $L^{cls}_{parts}$, $L^{rank}$, and $\cdot L^{reg}$ are the classification loss on image, the classification loss on discriminative parts, the ranking loss on keeping consistency of scores for parts, and the regularization loss of image representation, respectively. In testing, for the stable generalization performance in an ensemble learning, we combine multiple prediction outputs with equal weights as follows 
\begin{equation} \label{eqn:final_prediction}
    \bm{\hat{y}}^{(final)} = \sum^{S+1}_{s=1}{\bm{\hat{y}}^{(s)}},
\end{equation}
where the maximum entry in the vector $\bm{\hat{y}}^{(final)}$ corresponds to the class prediction. 
More importantly, most of the proposed components are only activated during training.
In other words, during testing, 
only the backbone branch (\ie, the bottom one in \cref{fig:architecture}) with curriculum learning is utilized to make predictions. 
In summary, computational cost of our P2P-Net is slightly greater than its original backbone (\eg ResNet) at inference.

\section{Experiments}\label{sec:experiment}
\subsection{Datasets and Settings}

We evaluate the proposed method on widely used benchmarks for fine-grained visual classification, including Caltech-UCSD Birds (CUB) \cite{birds2011}, the Stanford Cars (CAR) \cite{cars2013} and the FGVC Aircraft (AIR) \cite{aircraft2013}.
The \textbf{CUB} is a dataset with 11,788 images from 200 bird species. Data split is fixed with 5,994 images for training and 5,794 images for testing.
The \textbf{CAR} dataset consists of 16,185 images of 196 classes of cars, whose data is divided into 8,144 training images and 8,041 testing images, where each class has been split roughly in a 50-50 split. 
The \textbf{AIR} benchmark contains 10,000 images of 100 aircraft variants, among which only 3,333 images for testing.
%
%
We follow the same image and part size as recent work \cite{NTSNET2018}, \ie, raw images are first resized into $550 \times 550$ followed by random horizontal flipping and cropping (center cropping used in testing) into $448 \times 448$.
In the experiments, only class labels are available without any additional prior knowledge.

We evaluate the P2P-Net on the ResNet34 and the ResNet50 backbones, with weights pre-trained on the ImageNet dataset \cite{imagenet2009}.
Models are trained for 300 epochs with a mini-batch size of 16. 
Learning rate follows the cosine annealing schedule \cite{cos2016} with initial learning rate setting as $0.002$. 
Particularly, the learning rate of the backbone is set as one-tenth as other layers to make training more stable. 
The number of parts ($N$) and intermediate feature maps ($S$) are 4 and 3 as default values.
The loss weight $\beta$ in \cref{eqn:total_loss} is empirically set as 0.1.
For the smoothing factors, we simply set $\{\alpha^{(s)}\}$ as $\{0.7, 0.8, 0.9, 1.0\}$ in an ascending order for the fact that larger $\alpha$ indicates higher confidence of class predictions.

\subsection{Comparison with State-of-the-arts} \label{sec:peer_comparison}
\begin{table}[t]
    \centering
    \begin{tabular}{|l|c|c|c|c|}
    \hline
        \multirow{2}{*}{Method} & \multirow{2}{*}{Backbone} & \multicolumn{3}{|c|}{Accuracy (\%)} \\
        \cline{3-5}
         & & CUB & CAR & AIR \\
        \hline
        B-CNN \cite{BCNN2015} & VGG & 84.1 & 91.3 & 84.1 \\
        \hline
        RA-CNN \cite{RACNN2017} & \multirow{2}{*}{VGG19} & 85.3 & 92.5 & 88.2 \\
        MA-CNN \cite{MACNN2017} & & 86.5 & 92.8 & 89.9 \\
        \hline
        FCAN \cite{FCAN2016} & \multirow{13}{*}{ResNet50} & 84.7 & 93.1 & - \\
        MAMC \cite{MAMC2018} & & 86.3 & 93.0 & - \\
        DFL-CNN \cite{DFCNN2018} & & 87.4 & 93.1 & 91.7 \\
        NTS-Net \cite{NTSNET2018} & & 87.5 & 93.9 & 91.4 \\
        DCL \cite{DCL2019} & & 87.8 & 94.5 & 93.0 \\
        TASN \cite{TASN2019} & & 87.9 & 93.8 & - \\
        Cross-X \cite{CROSSX2019} & & 87.7 & 94.6 & 92.6 \\
        S3N \cite{S3N2019} & & 88.5 & 94.7 & 92.8 \\
        LIO \cite{LIO2020} & & 88.0 & 94.5 & 92.7 \\
        BNT \cite{BNT2020} & & 88.1 & 94.6 & 92.4 \\
        ASD \cite{ASD2020} & & 88.6 & 94.9 & 93.5 \\
        DF-GMM \cite{DFGMM2020} & & 88.8 & 94.8 & 93.8 \\
        PMG \cite{PMG2020} & & 89.6 & 95.1 & 93.4 \\
        \hline
        API-Net \cite{api-net2020} & ResNet101 & 88.6 & 94.9 & 93.4 \\
        \hline
        API-Net \cite{api-net2020} & DenseNet-161 & 90.0 & 95.3 & 93.9 \\
        \hline
        P2P-Net (ours) & ResNet34 & 89.5 & 94.9 & 92.6  \\
        \hline
        P2P-Net (ours) & ResNet50 & \textbf{90.2} & \textbf{95.4} & \textbf{94.2} \\
        \hline
    \end{tabular}
    \caption{Comparison with the state-of-the-art methods.}
    \label{tab:sota_performance}\vspace{-0.3cm}
\end{table}

Experimental results about comparative evaluation with recent fine-grained classifiers on the CUB\_200\_2011, the Stanford Cars, and the FGVC-Aircraft are illustrated in \cref{tab:sota_performance}. 
As the table shows, with the same backbone ResNet50, the proposed P2P-Net outperforms the state-of-the-art methods on three widely-used benchmarks by more than 0.4\% at average (+0.6\% on CUB, +0.3\% on CAR, and +0.4\% on AIR), whose performance gain can be credited to our proposed components.
What's more, our method also can achieve competitive performance to existing methods even with a shallow backbone -- ResNet34. 


\subsection{Ablation Studies} \label{sec:ablation_study}
\begin{table}[t]
    \centering
    \begin{tabular}{|l|c|c|c|}
    \hline
        \multirow{2}{*}{Method} & \multicolumn{3}{|c|}{Accuracy (\%)}  \\
        \cline{2-4}
        & CUB & CAR & AIR\\
        \hline
        (a) Baseline & 85.5 & 92.7 & 90.3 \\
        \hline
        (b) Baseline+CS & 88.4 & 94.9 & 93.8 \\
        \hline
        (c) Baseline+FR (w/o UPA) & 89.0 & 94.8 & 92.0 \\
        \hline
        (d) Baseline+FR (w/ UPA) & 89.0 & 95.0 & 92.5 \\
        \hline 
        (e) Baseline+FC & 88.4 & 94.7 & 93.8 \\
        \hline
        (f) Baseline+CS+FR (w/o UPA) & 90.0 & 95.0 & 93.9 \\
        \hline
        (g) Baseline+CS+FR (w/ UPA) & \textbf{90.2} & \textbf{95.4} & \textbf{94.2} \\
        \hline
    \end{tabular}
    \caption{Ablation study on the proposed components with the baseline ResNet50. CS: curriculum supervision, FR: feature regularization, UPA: unsupervised part alignment, FC: feature concatenation.}
    \label{tab:ablation_acc} \vspace{-0.3cm}
\end{table}

We conduct ablation studies to validate the effectiveness on the key components of our method, including curriculum supervision, part-to-pose feature regularization, and unsupervised part alignment respectively. 
The baseline method consists of a backbone ResNet50 and a MLP. 
Based on the baseline, we evaluated different combinations of the proposed components,
whose results are presented in \cref{tab:ablation_acc}.

\vspace{0.1cm}\noindent \textbf{Curriculum Supervision (CS) --} In comparison with (a) and (b), the CS module leads to large performance gain varying from +2.1\% to +3.5\% in classification accuracy on three datasets. 
Such results can be explained in two aspects: 1) by making predictions at different depths on the network, complementary information of multi-granularity can be fused to capture discriminative object features; 2) by connecting more layers to the output, parameters in shallower layers become easier to optimize. 
We further conduct contrastive experiments on the CUB between using $\{\alpha^{(s)}\}$ and using one-hot labels (no label smoothing). The accuracy of different predictions of the former one is $\{81.7, 87.3, 85.8, 87.8, 88.4\}$, while the latter one's is $\{82.1, 87.0, 85.6, 87.3, 88.2\}$. 
Such a result proves the proposed smoothing scheme's effectiveness.

\vspace{0.1cm}\noindent \textbf{Feature Regularization (FR) --} Comparing variants (c-d) with (a), the FR module provides significant improvement on performance, while it shows smaller gain compared to the CS component on the AIR.
Since 
a global contour is an important cue for classifying big objects like aircrafts, when using local distinctive information to regularize the object representation learning, the result may not be better.
To verify our motivation, we report comparative evaluation with an additional competitor (e), whose network structure is almost the same as methods (c-d).
The mere difference lies in that, in (e) we use concatenation of features of local parts and a global image for classification, rather than the proposed feature regularization.
As expected, (e) outperforms (c-d) on the AIR but is slightly worse on other datasets.
Our explanation is that the global contour information of aircraft classes is well preserved when using feature concatenation, instead of regularizing image feature with relatively small parts.
It is worth emphasizing again that, using FR is more computationally efficient owing to discarding the part-to-pose regularization branch during testing.

\begin{table}[t]
    \centering
    \begin{tabular}{|l|c|c|c|}
    \hline
    \multirow{2}{*}{Method} & \multicolumn{3}{|c|}{RMSE}  \\
        \cline{2-4}
        & CUB & CAR & AIR\\
        \hline
        Baseline & 0.501 & 0.354 & 0.399 \\
        \hline
        Baseline+FC & 0.268 & 0.213 & 0.354 \\
        \hline
        Baseline+FR (w/ UPA) & \textbf{0.213} & \textbf{0.179} & \textbf{0.263} \\
    \hline
    \end{tabular}
    \caption{RMSE of learned representations of different methods.}
    \label{tab:representation_compactness}\vspace{-0.3cm}
\end{table}

\vspace{0.1cm}\noindent \textbf{Unsupervised Parts Alignment (UPA) --} Lastly, the method integrating FR with UPA outperforms its competitor without using UPA (See methods (c-d) and (f-g) in the table). 
These results demonstrate that the effectiveness of our unsupervised graph matching for part alignment, which can be a supplement to the part-based feature regularization. 
Besides, we also measure the compactness of image representations.
In details, for each class, we compute root mean square error (RMSE) of learned representations and reported the average in \cref{tab:representation_compactness}, where P2P-Net (\ie, the bottom one) achieves the least value.
The second method in the table can be viewed as a typical feature concatenation of the conventional part-based methods.
The result shows that the object representations learned with FR and UPA are more concentrated around their class centers in feature space.
As we know that, objects of the same class are of varied poses, the result also demonstrates that our method are able to align detected discriminative parts to narrow the intra-class variance caused by pose changes.

\vspace{0.1cm}\noindent \textbf{Size of Parts --} Result of using different parts number ($N$) on the CUB are shown in Tab.~\ref{tab:topn}. Larger values may not bring a significant improvement, so $N=4$ is preferred. 
\begin{table}[htbp]
    \centering
    \begin{tabular}{|l|c|c|c|c|c|}
    \hline
        Parts number ($N$) & 2 & 3 & 4 & 5 & 6 \\
    \hline
        CUB & 88.1 & 89.9 & \textbf{90.2} & \textbf{90.2} & 90.0 \\
    \hline
    \end{tabular}
    \caption{Influence of the number of discriminative parts.}
    \label{tab:topn}\vspace{-0.3cm}
\end{table}

\subsection{Visualizations} \label{sec:visualization}

\begin{figure}[t]
    \centering
    \includegraphics[width=0.95\linewidth]{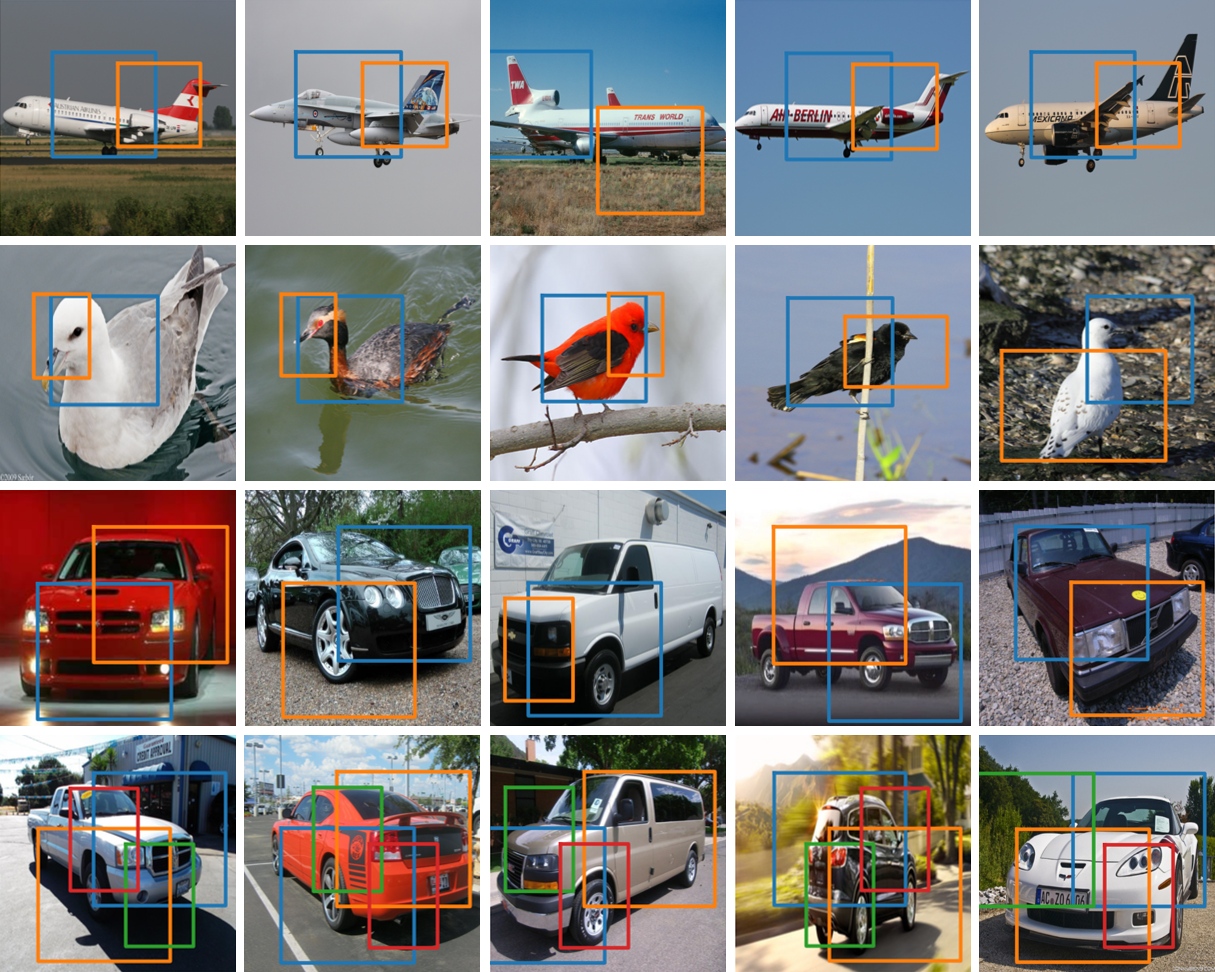}
    \caption{Discriminative parts detected by our P2P-Net.}
    \label{fig:parts_location}\vspace{-0.3cm}
\end{figure}

\vspace{0.1cm}\noindent \textbf{Parts Location --} The discriminative parts located by our P2P-Net are visualized in \cref{fig:parts_location}.
Only the top $2$ parts' bounding boxes are shown from the 1st to the 3rd row, while top $4$ salient parts are highlighted in the 4th row.
As shown in the figure, the top $2$ discriminative parts share visual similarity on localized parts, although they are detected in weakly-supervised learning manner. 
Specifically, for aircraft samples, salient regions tend to locate in the body and tail; for bird breeds, they usually concern on birds' head and body; and for cars, the front and body of vehicles contain discriminative details.
This phenomenon is an important prerequisite for the proposed graph matching method.

\begin{figure}[t]
    \centering
    \includegraphics[width=0.92\linewidth]{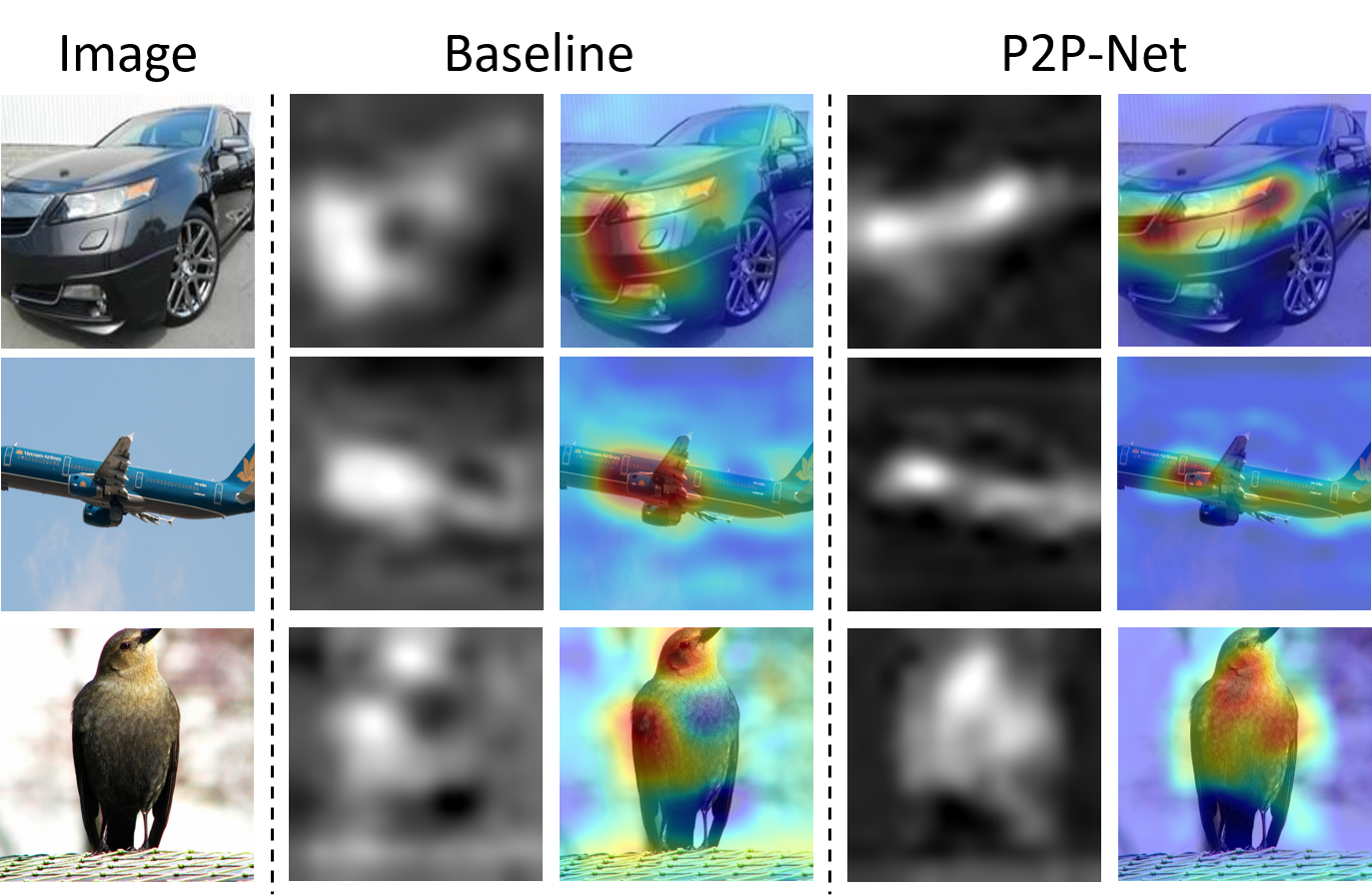}
    \caption{Class activation maps of some test samples.}
    \label{fig:cam} \vspace{-0.3cm}
\end{figure}
\vspace{0.1cm}\noindent \textbf{Class Activation Map --} We also apply Grad-CAM \cite{gradcam2017} to the last convolutional layer to show intuitive visualization. 
Both gray scale and merged color activation images are illustrated in \cref{fig:cam}.
Compared to the baseline, our P2P-Net has less activation on the background and is more concentrated on the discriminative regions of objects.
Evidently, it is verified that our P2P-Net is able to extract discriminative information from distinctive regions and eliminate the impact of noisy backgrounds.

\begin{figure}[t]
    \centering
    \includegraphics[width=0.92\linewidth]{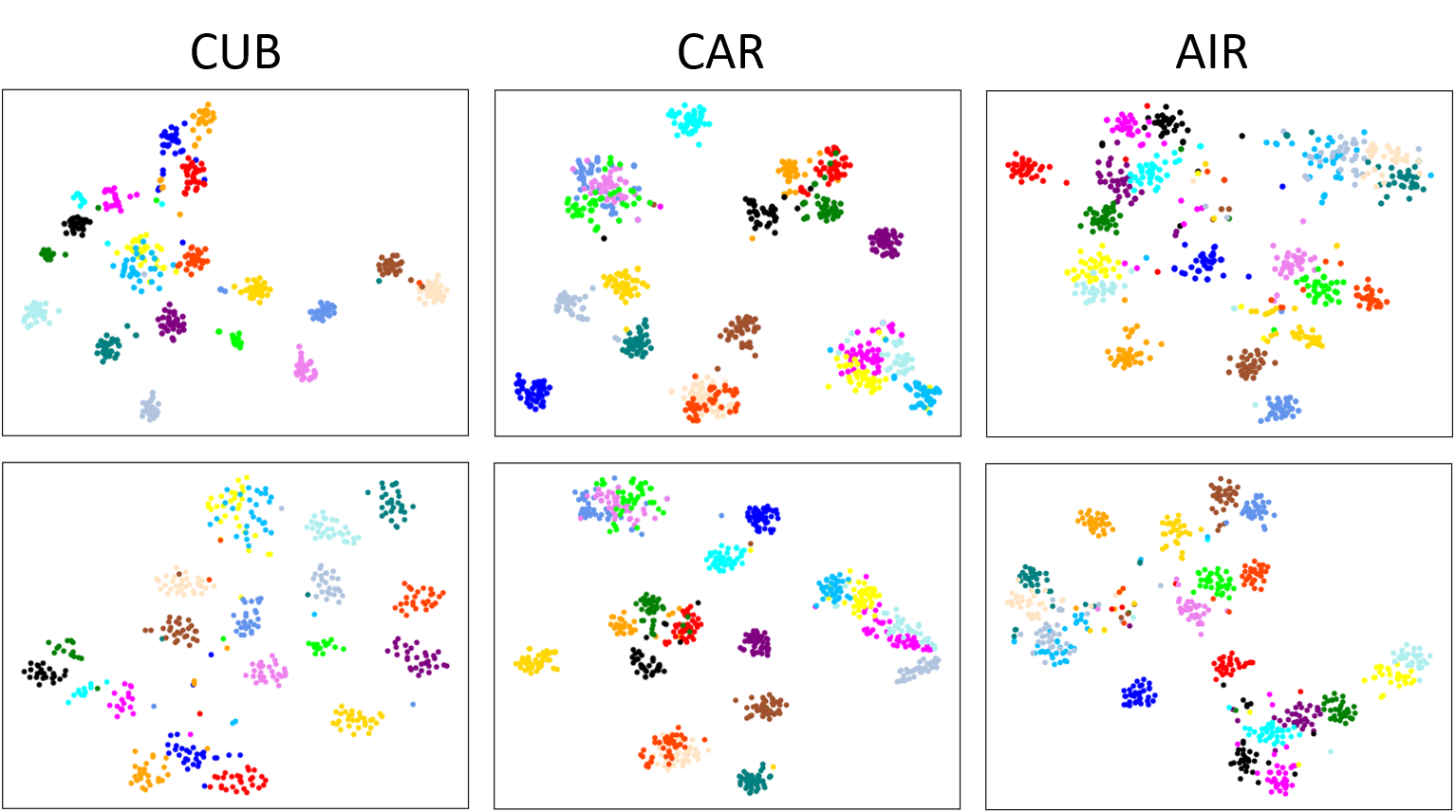}
    \caption{A T-SNE plot of learned representations on three datasets. First row: the baseline model; second row: baseline+FR (w/ UPA).
    }
    \label{fig:tsne_feature} \vspace{-0.5cm}
\end{figure}
\vspace{0.1cm}\noindent \textbf{Feature Visualization --} In \cref{fig:tsne_feature}, we draw the t-SNE scatter plot from the learned high dimensional features. 
On each dataset, we randomly selected twenty classes for visualization.
As the t-SNE plot shows, after applying parts' feature regularization, image representations exhibit higher intra-classes variations and favor for enlarging inter-class margins, especially on the CUB and AIR datasets.

\section{Conclusion}\label{sec:conclusion}

This paper proposes a novel feature regularization scheme about learning pose-insensitive discriminative representations for fine-grained classification.
Our P2P-Net not only utilizes the located discriminative parts to promote discrimination of image representations, but also introduces graph matching for part alignment for robustness against pose variations.
Besides, a curriculum supervision strategy is verified to further boost the performance without paying any price of extra annotations.
Extensive experiments on well-known benchmarks against the state-of-the-art methods and ablation analysis have demonstrated the effectiveness of our method.
In addition, the contributed scheme might be under adversarial attacks, causing total failure of the whole perception system, which encourages researchers and safety engineers to mitigate these risks.

\section*{Acknowledgements}
This work is supported by the China Postdoctoral Science Foundation (2021M691682), the National Natural Science Foundation of China (61902131, 62072188, U20B2052), the Program for Guangdong Introducing Innovative and Entrepreneurial Teams (2017ZT07X183), and the Project of Peng Cheng Laboratory (PCL2021A07).

{\small
\bibliographystyle{ieee_fullname}
\bibliography{Paper.bbl}
}

\end{document}